\documentclass[conference]{IEEEtran}
\usepackage{times}

\usepackage[numbers]{natbib}
\usepackage{multicol}
\usepackage[bookmarks=true]{hyperref}
\usepackage{graphicx}
\graphicspath{ {./images/} }
\usepackage{fancyvrb,fvextra}
\usepackage{float}
\floatstyle{ruled}
\newfloat{code}{thp}{lop}
\newfloat{result}{tbp}{lop}
\floatname{code}{Code}
\floatname{result}{Result}

\begin{document}

\title{Agent 3, change your route: possible conversation between a human manager and UAM Air Traffic Management (UATM)}


\author{\authorblockN{Jeongseok Kim}
\authorblockA{AIX\\
SK Telecom\\
Seoul, Korea\\
Email: jeongseok.kim@sk.com}
\and
\authorblockN{Kangjin Kim}
\authorblockA{Department of Drone Systems\\
Chodang University\\
Jeollanam-do, Korea\\
Email: kangjinkim@cdu.ac.kr}}


%

\maketitle

\begin{abstract}
This work in progress paper provides an example to show a detouring procedure through knowledge representation and reasoning. When a human manager requests a detouring, this should affect the related agents. Through non-monotonic reasoning process, we verify each step to be proceeded and provide all the successful connections of the reasoning. Following this progress and continuing this idea development, we expect that this simulated scenario can be a guideline to build the traffic management system in real. After a brief introduction including related works, we provide our problem formulation, primary work, discussion, and conclusions.
\end{abstract}

\IEEEpeerreviewmaketitle

\section{Introduction}

In the recent advancement of public transportation, Urban Air Mobility (UAM) has gained significant attention. However, its unique challenges – such as low-altitude flights interwoven with ground-level life spaces, high-density air traffic, and an anticipated transition to fully autonomous operations by 2035\cite{KUAMConops10} – make the existing aviation systems inadequate for managing this emerging paradigm.

This research navigates through these complexities using UAM Air Traffic Management (UATM). A graph-based model is used to represent the UAM airway network, where each vertex signifies a `vertiport' (vertical airport) and each edge denotes a corridor between two adjacent vertiports. A human traffic manager is allocated to each vertiport, who is responsible for supervising the landing and take-off procedures and communicating any traffic issues to the UATM.
We demonstrate through a practical scenario in this paper, how detouring orders can be delivered to relevant agents leading to a change in their current paths. It's crucial to note that the coverage range of each UATM is not determined by its proximity to a vertiport but is independently established by the UATM itself. This autonomous setting becomes increasingly vital considering the limitations of communication ranges. In some situations, a UATM may have to relay its orders to certain agents through the UATM Network\cite{JS2022tbo}.
Through this study, we aim to unravel the operational complexities in the fast-evolving UAM sector and contribute to a safer and more efficient urban air traffic management system in the future.

\subsection{Related Works}

The UAM is a fast-growing industry and a rapidly evolving field of research. The papers \cite{Reiche2018, Garrow2021, KUAMConops10, Marzouk2022} provide a comprehensive overview of the technology, regulatory landscape,  potential benefits, and challenges of UAM.

This paper \cite{Schuchardt2023} revisits German Aerospace Center (DLR) initiatives on urban air traffic management and informs ongoing air traffic management research. The authors in \cite{Kim2022} proposes a risk assessment model to evaluate collision risks between vehicles and obstacles, based on past studies. The paper \cite{PintoNeto2023} provides an in-depth review of contemporary deep learning (DL) solutions for air traffic management, as well as an extensive list of outstanding challenges.


The aforementioned efforts offer air traffic management-based collision prevention options, however none of them persuade various stakeholders. In UATM, multiple stakeholders, mass agent actions, and catastrophic accident monitoring can complicate the system. This paper attempts non-monotonic reasoning to demonstrate route detouring, which is common in these complex systems.


Along this line, explainable AI (XAI) is a new research subject, and the paper \cite{BorregoDiaz2022} is a good reference. It thoroughly introduces explainable AI. They offer epistemological modeling, based on knowledge representation and reasoning, and believe this validates complicated system decision-making.

\subsection{Contributions}

We show a detour scenario as an example that can be frequently situated in UAM settings. This suggests a grounding in the necessary procedure for the detour request. 

\subsection{Outlines}
In the following section, we address the problem formulation, continue the primary solution, followed by possible discussions, and then conclude this paper.

\section{Problem Formulation}

\begin{figure}
	\centering
	\includegraphics[width=0.85\linewidth]{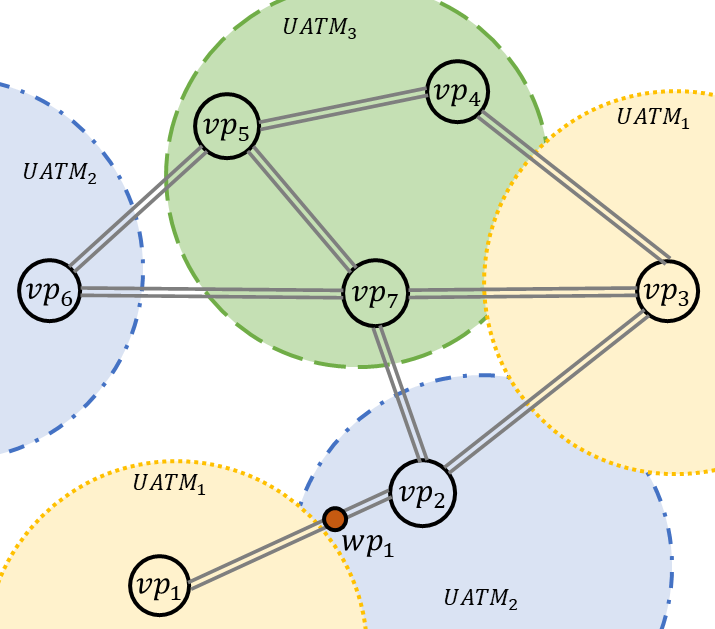}
	\caption{UATM Network, which consists of seven vertiports from $vp_1$ to $vp_7$, bidirectionally connecting corridors between adjacent vertiports, three UATMs of $UATM_1$, $UATM_2$, and $UATM_3$, and their coverage represented with outer circles.}
	\label{fig:uatm_network}
\end{figure}

Consider a network of vertiports, with some of these vertiports being adjacent and interconnected to form a corridor, as depicted in Fig.~\ref{fig:uatm_network}. Here, each vertex symbolizes a vertiport, and the edges linking them represent the corridors. Surrounding these vertiports are expansive circles, which indicate the communication coverage of the UAM Air Traffic Management (UATM) systems. In this network, agents journey from one vertiport to another via these corridors. A UATM can directly communicate with any agent within its coverage range. For the purposes of this study, we posit a communication relaying protocol enabling UATMs to relay messages through other UATMs - a system we refer to as the `UATM network'. However, in cases where the agent moves beyond the coverage of this UATM network, direct communication with the agent becomes impossible.

An agent's trip has the following order: First, it is located in a vertiport, ready to take off. This agent requests permission to take off from the traffic manager in the vertiport. As soon as he or she authorizes the agent to perform the desired action, it begins to ascend into the airspace.

We assume that the destination is already configured and that the route is also computed before flying. Visiting a sequence of corridors to reach the destination vertiport, the agent communicates UATMs in order to update its status, such as current velocity, GPS coordinates, and other necessary information. Then, UATMs can monitor the agent, considering the entire traffic situation.


The agent asks the vertiport traffic management to land when it gets close. The traffic manager guides it to the vertipad after confirmation.


This traffic system runs automatically without a human asistant. Each vertiport has a human traffic manager. They monitor traffic and the vertiport environment, and also interact with legacy traffic systems. Thus, human managers make the system more adaptable.

\subsection{A Detouring Scenario}
While a human manager in the vertiport \#3 ($vp_3$ in short) monitored landing and taking off for a while, he or she found some delays. If these delays are accumulated, and propagated to the corridors, it is anticipated that some collisions will occur. Hence, he/she reported this situation to the UATM (here, $UATM_1$). Particularly, the corridor from $vp_2$ to $vp_3$ is condensed so that it is required for agents to avoid using that corridor. This leads to detouring the route of agents that are heading to $vp_3$. The alternative route is a sequence of vertiports $vp_1$, $vp_2$, $vp_7$, and $vp_3$.

From the $UATM_1$'s point of view, this request from the manager requires somewhat extended work. Once it receives a detour request from $vp_3$ manager, it has to find all upcoming agents for $vp_3$. Then, it samples agents heading to $vp_2$, and it sends them a new route, which is a sequence of vertiports $vp_1$, $vp_2$, $vp_7$, and $vp_3$. In addition, $UATM_1$ should check if some agents are out of $UATM_1$'s coverage. If such agents exist, it should find the proper UATMs by querying the entire UATM network. For example, suppose that there is an agent \#3 at waypoint 1 ($wp_1$ in short). Then, $UATM_2$ will respond to the query through the network. In this case, $UATM_1$ will ask $UATM_2$ to deliver the new route to agent \#3. After asking for this detour message to be relayed, $UATM_1$ will wait for the response from $UATM_2$ about the fact that agent \#3 updated its plan. Once it finds that its plan is renewed, $UATM_1$ can respond back to $vp_3$ manager.

\section{Solution}
In this section, we declare the problem in Answer Set Programming. While developing the scenario with several small questions, we provide non-monotonic reasoning. Following the result, we can check if a given mission is satisfiable. If it is not, we can find out which factors will make them fail.
\subsection{A Common Setting}
For the entire solution, there is a common setting. Due to the page constraint, we place the Code~\ref{code:common_settings} as an appendix. We will use this code before starting any other code.
\subsection{Basic Queries}

\subsubsection{Find all covered agents by $UATM_1$}
We here find all agents that are covered by $UATM_1$. It is worth noting that in Code~\ref{code:query_03} we intentionally omitted codes for agents' moving through the step. The code has two major heads of rules. One is \texttt{covered\_agent} and another is \texttt{covered\_by\_uatm1}. All agents in \texttt{covered\_agent} are matched with their UATM, and then in the second head \texttt{covered\_by\_uatm1} we can sample the agents that are only covered by $UATM_1$.
As shown in Result~\ref{res:query_03}, agents \#1, \#2, \#4 and \#5 are covered by $UATM_1$.

\begin{code}
	\begin{Verbatim}[breaklines,fontsize=\footnotesize]
plan(3, 1, 1, 2).
plan(3, 1, 2, 3).

covered_agent(A, TM) :- loc(A, T, U, V, WP), covered_wp(U, V, TM, WP).
covered_by_uatm1(A) :- covered_agent(A, 1).

#show covered_by_uatm1/1.
	\end{Verbatim}
	\caption{Find all agents that are covered by $UATM_1$}
	\label{code:query_03}
\end{code}

\begin{result}
	\begin{Verbatim}[breaklines,fontsize=\footnotesize]
$ clingo conv04.lp
clingo version 5.6.2
Reading from conv04.lp
Solving...
Answer: 1
covered_by_uatm1(1) covered_by_uatm1(2) covered_by_uatm1(4) covered_by_uatm1(5) loc(1,1,1,2,3) loc(2,1,1,2,7) loc(3,1,1,2,17) loc(4,1,1,2,12) loc(5,1,1,2,15) loc(6,1,1,2,19)
SATISFIABLE

Models       : 1
Calls        : 1
Time         : 0.003s (Solving: 0.00s 1st Model: 0.00s Unsat: 0.00s)
CPU Time     : 0.000s
	\end{Verbatim}
	\caption{All agents covered by $UATM_1$}
	\label{res:query_03}
\end{result}

\subsubsection{Change the route for covered, $vp_3$ heading agents}
This is to change the route for the following agents:
\begin{itemize}
	\item they are within the $UATM_1$'s coverage,
	\item their original plan is supposed to pass the exclusive edge,
	\item their target is the $vp_3$, and
	\item they are on the edge between $vp_1$ and $vp_2$ now, which means that they are before visiting the exclusive edge.
\end{itemize}

We assume that there are six related agents for this query.
In the beginning, for these six agents, we provide the initial plan, \texttt{plan}, which is a sequence of vertiports $vp_1$, $vp_2$, and $vp_3$.
This is presented in the first two lines in Code~\ref{code:query_05}.
We then give a new plan, \texttt{new\_plan}, which is a sequence of vertiports $vp_1$, $vp_2$, $vp_7$, and $vp_3$.
The first value 2 of \texttt{new\_plan} does not indicate any specific agent id, but directs a step number.
It means that at step 2, the new plan should be applied to some agents.
After covered agents are found, the head \texttt{detour\_request} checks if their target is $vp_3$ and they are before visiting the corridor between $vp_2$ and $vp_3$.
Then, for the particular agent, \texttt{detour\_request} is made for the one time step ahead.
Once the time is passed by one step, agents having \texttt{detour\_request} change their plan with the \texttt{new\_plan}.
From Result~\ref{res:query_05}, we can see four agents have the detour request, and they all changed their route.
Since it is \texttt{SATISFIABLE}, we know that \texttt{detour\_request} is applied.

\begin{code}
	\begin{Verbatim}[breaklines,fontsize=\footnotesize]
plan(A, 1, 1, 2) :- agent(A), 1 <= A, A <= 6.
plan(A, 1, 2, 3) :- agent(A), 1 <= A, A <= 6.

% we assume that every plan is acyclic.
source(A, U) :- agent(A), plan(A, 1, U, V), not plan(A, 1, _, U).
target(A, V) :- agent(A), plan(A, 1, U, V), not plan(A, 1, V, _).

new_plan(2, 1, 2).
new_plan(2, 2, 7).
new_plan(2, 7, 3).

plan(A, T+1, U, V) :- plan(A, T, U, V), step(T+1), not detour_request(A, T+1).
plan(A, T+1, U1, V1) :- plan(A, T, U, V), step(T+1), new_plan(T+1, U1, V1), detour_request(A, T+1).

covered_agent(A, TM) :- loc(A, T, U, V, WP), covered_wp(U, V, TM, WP).
covered_by_uatm1(A) :- covered_agent(A, 1).

detour_request(A, T+1) :- covered_by_uatm1(A), plan(A, T, U, V), plan(A, T, 2, 3), target(A, 3), edge_range(2, 3, P), not loc(A, T, 2, 3, P), not step(T-1).

change_route(A, T) :- new_plan(T, U, V), plan(A, T, U, V), detour_request(A, T).
:- not change_route(A, T), new_plan(T, U, V), detour_request(A, T).

#show detour_request/2.
#show change_route/2.
	\end{Verbatim}
	\caption{Change the route for covered, $vp_3$ heading agents}
	\label{code:query_05}
\end{code}

\begin{result}
	\begin{Verbatim}[breaklines,fontsize=\footnotesize]
$ clingo conv03c.lp
clingo version 5.6.2
Reading from conv03c.lp
Solving...
Answer: 1
detour_request(5,2) detour_request(4,2) detour_request(2,2) detour_request(1,2) change_route(5,2) change_route(4,2) change_route(2,2) change_route(1,2)
SATISFIABLE

Models       : 1
Calls        : 1
Time         : 0.004s (Solving: 0.00s 1st Model: 0.00s Unsat: 0.00s)
CPU Time     : 0.000s
	\end{Verbatim}
	\caption{All covered agents' plans are renewed}
	\label{res:query_05}
\end{result}

\subsection{Advanced Queries}
\subsubsection{Find all the unreachable with $UATM_1$, but $vp_3$ heading agents}
As we know from Code~\ref{code:query_03}, we can find covered agents by $UATM_1$ through the head of the rules \texttt{covered\_by\_uatm1}. From this head, we can sample the covered agents.
Comparing these agents with all the agents located in the corridor between $vp_1$ and $vp_2$, we can find uncovered agents. The head of the rule \texttt{uncovered\_by\_uatm1} has these remaining agents in Code~\ref{code:query_06}.
As shown in Result~\ref{res:query_06}, there are two uncovered agents: agent \#3 and agent \#6.
\begin{code}
	\begin{Verbatim}[breaklines,fontsize=\footnotesize]
plan(A, 1, 1, 2) :- agent(A), 1 <= A, A <= 6.
plan(A, 1, 2, 3) :- agent(A), 1 <= A, A <= 6.

% we assume that every plan is acyclic.
source(A, U) :- agent(A), plan(A, 1, U, V), not plan(A, 1, _, U).
target(A, V) :- agent(A), plan(A, 1, U, V), not plan(A, 1, V, _).

covered_agent(A, TM) :- loc(A, T, U, V, WP), covered_wp(U, V, TM, WP).
uncovered_by_uatm1(A) :- not covered_agent(A, 1), loc(A, T, 1, 2, _), plan(A, T, 2, 3), target(A, 3).

#show uncovered_by_uatm1/1.
	\end{Verbatim}
	\caption{Find all unreachable agents by $UATM_1$}
	\label{code:query_06}
\end{code}

\begin{result}
	\begin{Verbatim}[breaklines,fontsize=\footnotesize]
$ clingo conv05c.lp
clingo version 5.6.2
Reading from conv05c.lp
Solving...
Answer: 1
uncovered_by_uatm1(3) uncovered_by_uatm1(6)
SATISFIABLE

Models       : 1
Calls        : 1
Time         : 0.002s (Solving: 0.00s 1st Model: 0.00s Unsat: 0.00s)
CPU Time     : 0.000s
	\end{Verbatim}
	\caption{All unreachable agents by $UATM_1$}
	\label{res:query_06}
\end{result}

\subsubsection{Change the route for all $vp_3$ heading agents}
Code~\ref{code:query_07} has two different heads of the rules for \texttt{detour\_request}. One is for covered agents by $UATM_1$ and another is for unreachable agents by $UATM_1$.
As shown in Result~\ref{res:query_07}, all covered and uncovered agents are specified, followed by their detour requests, and then finally their route changes are presented at time step 2.

\begin{code}
	\begin{Verbatim}[breaklines,fontsize=\footnotesize]
plan(A, 1, 1, 2) :- agent(A), 1 <= A, A <= 6.
plan(A, 1, 2, 3) :- agent(A), 1 <= A, A <= 6.

% we assume that every plan is acyclic.
source(A, U) :- agent(A), plan(A, 1, U, V), not plan(A, 1, _, U).
target(A, V) :- agent(A), plan(A, 1, U, V), not plan(A, 1, V, _).

new_plan(2, 1, 2).
new_plan(2, 2, 7).
new_plan(2, 7, 3).

plan(A, T+1, U, V) :- plan(A, T, U, V), step(T+1), not detour_request(A, T+1).
plan(A, T+1, U1, V1) :- plan(A, T, U, V), step(T+1), new_plan(T+1, U1, V1), detour_request(A, T+1).

covered_agent(A, TM) :- loc(A, T, U, V, WP), covered_wp(U, V, TM, WP).
covered_by_uatm1(A) :- covered_agent(A, 1).
uncovered_by_uatm1(A) :- not covered_agent(A, 1), loc(A, T, 1, 2, _), plan(A, T, 2, 3), target(A, 3).
covered(A, T, TM) :- loc(A, T, U, V, WP), uncovered_by_uatm1(A), covered_wp(U, V, TM, WP).

detour_request(A, T+1) :- covered_by_uatm1(A), plan(A, T, U, V), plan(A, T, 2, 3), target(A, 3), edge_range(2, 3, P), not loc(A, T, 2, 3, P), not step(T-1).
detour_request(A, T+1) :- covered(A, T, TM), plan(A, T, U, V), plan(A, T, 2, 3), target(A, 3), edge_range(2, 3, P), not loc(A, T, 2, 3, P), not step(T-1).

change_route(A, T) :- new_plan(T, U, V), plan(A, T, U, V), detour_request(A, T).
:- not change_route(A, T), new_plan(T, U, V), detour_request(A, T).

#show covered_by_uatm1/1.
#show uncovered_by_uatm1/1.
#show detour_request/2.
#show change_route/2.
	\end{Verbatim}
	\caption{Change the route for all $vp_3$ heading agents}
	\label{code:query_07}
\end{code}

\begin{result}
	\begin{Verbatim}[breaklines,fontsize=\footnotesize]
$ clingo conv07c.lp
clingo version 5.6.2
Reading from conv07c.lp
Solving...
Answer: 1
covered_by_uatm1(1) covered_by_uatm1(2) covered_by_uatm1(4) covered_by_uatm1(5) detour_request(6,2) detour_request(5,2) detour_request(4,2) detour_request(3,2) detour_request(2,2) detour_request(1,2) uncovered_by_uatm1(3) uncovered_by_uatm1(6) change_route(6,2) change_route(5,2) change_route(4,2) change_route(3,2) change_route(2,2) change_route(1,2)
SATISFIABLE

Models       : 1
Calls        : 1
Time         : 0.007s (Solving: 0.00s 1st Model: 0.00s Unsat: 0.00s)
CPU Time     : 0.000s
	\end{Verbatim}
	\caption{All $vp_3$ heading agents' plans are renewed}
	\label{res:query_07}
\end{result}

\section{Discussion}
\subsection{Current Progress}
This is our initial step into this field of research. We realized that language between human traffic managers and the traffic system can be used intuitively to interact with each other. In addition, utilizing the KRR, each stackholder can trace any event in the system with confidence.

\subsection{Future Directions}

We are adding scenarios to our job. We mostly want to add agent movement. Second, reactively simulate the scenarios. Third, theoretical contributions. Finally, we want to incorporate a large language model. This should help managers interact with the system.

\subsection{Final Deliverable}
We want to have a comprehensive air traffic management system. This will be intuitively manageable, and each order will be easily traceable. Once an event occurs, we will find the cause of the event efficiently.

\section{Conclusions} 
\label{sec:conclusion}

We have outlined a scenario related to UATM. By employing KRR as a tool to articulate system explanations, we have instantiated KRR within the framework of Answer Set Programming. This paper then explores our current progress, future aspirations, and the ultimate objectives we aim to achieve in this context.

\section*{Acknowledgments}
This work is supported by the Korea Agency for Infrastructure Technology Advancement(KAIA) grant funded by the Ministry of Land, Infrastructure and Transport (Grant RS-2022-00143965).

\bibliographystyle{plainnat}
\bibliography{references}

\begin{appendices}

\section{Common Setting}
\begin{code}
	\begin{Verbatim}[breaklines,fontsize=\footnotesize]
uatm(1..3).
agent(1..20).

edge(1, 2).
edge(2, 3).
edge(2, 7).
edge(7, 3).

vp(1..7).

loc(1, 1, 1, 2, 3).
loc(2, 1, 1, 2, 7).
loc(3, 1, 1, 2, 17).
loc(4, 1, 1, 2, 12).
loc(5, 1, 1, 2, 15).
loc(6, 1, 1, 2, 19).

cover(1, 1).
cover(1, 3).
cover(2, 2).
cover(3, 7).

edge_range(1, 2, 1..20).
edge_range(2, 3, 1..13).
edge_range(2, 7, 1..22).

covered_wp(1, 2, 1, P) :- 1 <= P, P < 16, edge_range(1, 2, P).
covered_wp(1, 2, 2, P) :- 7 <= P, P <= 20, edge_range(1, 2, P).
covered_wp(2, 3, 1, P) :- 9 <= P, P <= 13, edge_range(2, 3, P).
covered_wp(2, 3, 2, P) :- 1 <= P, P < 9, edge_range(2, 3, P).
covered_wp(2, 7, 2, P) :- 1 <= P, P < 8, edge_range(2, 7, P).
covered_wp(2, 7, 3, P) :- 20 <= P, P <= 22, edge_range(2, 7, P).

step(1..3).
	\end{Verbatim}
	\caption{Common Setting}
	\label{code:common_settings}
\end{code}

\end{appendices}

\end{document}